\documentclass[10pt,twocolumn,letterpaper]{article}
\usepackage{iccv}
\usepackage{pifont}
\usepackage{times}
\usepackage{epsfig}
\usepackage{graphicx}
\usepackage{amsmath}
\usepackage{amssymb}
\usepackage{diagbox}
\usepackage{makecell}
\usepackage{booktabs}
\usepackage{indentfirst} 
\usepackage{amsmath, amssymb}
\usepackage{multirow}
\usepackage[accsupp]{axessibility}  

\usepackage[pagebackref=true,breaklinks=true,letterpaper=true,colorlinks,bookmarks=false]{hyperref}
\iccvfinalcopy 

\def\iccvPaperID{7818} 

\ificcvfinal\fi
\begin{document}

\title{Matching in the Dark: A Dataset for Matching Image Pairs of Low-light Scenes}

\author{Wenzheng Song$^{1}$ ~~~~Masanori Suganuma$^{1,2}$ ~~~~Xing Liu$^{1}$ \\ ~~~~Noriyuki Shimobayashi$^{1}$ ~~~~Daisuke Maruta$^{3}$ ~~~~Takayuki Okatani$^{1,2}$ \\
$^{1}$GSIS, Tohoku University ~~~~~~~~~$^{2}$RIKEN Center for AIP ~~~~~~~~~$^{3}$Socionext Inc.\\
{\tt\small \{song,suganuma,ryu,nshimobayashi,okatani\}@vision.is.tohoku.ac.jp}\\
{\tt\small {maruta.daisuke@socionext.com}}}


\maketitle
\ificcvfinal\thispagestyle{empty}\fi

\begin{abstract}
This paper considers matching images of low-light scenes, aiming to widen the frontier of SfM and visual SLAM applications. Recent image sensors can record the brightness of scenes with more than eight-bit precision, available in their RAW-format image. We are interested in making full use of such high-precision information to match extremely low-light scene images that conventional methods cannot handle. For extreme low-light scenes, even if some of their brightness information exists in the RAW format images' low bits, the standard raw image processing on cameras fails to utilize them properly. As was recently shown by Chen et al.\cite{chen2018learning}, CNNs can learn to produce images with a natural appearance from such RAW-format images. To consider if and how well we can utilize such information stored in RAW-format images for image matching, we have created a new dataset named MID (matching in the dark). Using it, we experimentally evaluated combinations of eight image-enhancing methods and eleven image matching methods consisting of classical/neural local descriptors and classical/neural initial point-matching methods. The results show the advantage of using the RAW-format images and the strengths and weaknesses of the above component methods. They also imply there is room for further research.  
\end{abstract}

\section{Introduction}
\label{section:1}
Structure-from-motion (SfM) \cite{sfm1, sfm2} and visual SLAM (simultaneous localization and mapping) \cite{slam1, slam2} have been used for real-world applications for a while. The mainstream methods use point correspondences between multiple views of a scene. They first detect keypoints and extract the descriptor of the local feature at each keypoint \cite{sift,mikolajczyk2004scale,akaze,orb}. They then find initial point correspondences between images and eliminate outliers from them, finally estimating the geometric parameters such as camera poses, etc.

SfM and visual SLAM have the potential to widen their application fields. One important target is the application to extremely low-light environments, such as outdoor scenes at night under moonlight or indoor scenes with insufficient illumination.  Making it possible to use SfM and visual SLAM in these environments is essential for real-world applications, such as autonomous vehicles that can operate at night. 

{\color{black} 
Owing to the advancement of image sensors, they can record incoming light with more than eight bits (e.g., 14 bits).
However, standard raw image processing employed on many cameras cannot make full use of the information existing in the lower bits of the sensor signal; it reduces mosaic artifacts on the sensor signal, adjusts the white balance and contrast, and then converts the processed signal into the standard format of eight-bit RGB images (we will refer to this raw image processing as RIP in this paper).
This limitation arguably comes from the requirement for versatility against all sorts of scenes with various lighting conditions in addition to reducing the number of bits. In extreme low-light scenes, even when some details of the scenes' brightness are stored in the low bits of their RAW signals, the standard RIP often yields mostly black images. The study of SID (see-in-the-dark) \cite{chen2018learning} well proves such limitation of the image pipeline, in which the authors show that a CNN can learn to convert such RAW-format images of dark scenes into brightened images with a natural appearance.


It is very likely that we can do the same with SfM and visual SLAM applied to low-light scenes, i.e., extracting the information present in the lower bits of the RAW signals to make SfM/visual SLAM work. The question is how to do this. It is noteworthy that the goal is not to generate natural looking bright images as SID does but to achieve the optimal performance for SfM and visual SLAM. 
}



There are potentially several directions to achieve the goal.
One is to develop a keypoint detector and a feature descriptor that work directly on the RAW-format images.
Even if keypoint detectors and descriptors are not good enough, it could be possible to attain the necessary level of matching performance by strengthening the subsequent steps in the pipeline. Recently, CNNs have been applied to these steps, leading to promising results, such as outlier removal in the initial correspondences \cite{moo2018learning,brachmann2019neural} and establishing initial matching \cite{superglue}. 
In parallel to these, the application of image enhancement methods for RAW-format low-light images in a pre-processing stage of image matching could be useful, e.g., SID \cite{chen2018learning} and others \cite{chen2019seeing,wei2020physics}. Methods for more general image restoration would be applied to the RAW-format images \cite{zhang2019residual,liu2019dual}.



As above, we can think of multiple different approaches to making SfM and visual SLAM methods applicable to low light environments. To promote further studies, we need a dataset to evaluate the above approaches in a multi-faceted fashion. Aiming at widening their application field toward lower-light scenes, it is necessary to examine how underexposed the image will be that each approach can deal with. There is currently no dataset that can be used for this purpose. Considering these, we create a dataset having the following features:
\begin{itemize}
\item 
To examine each method's limit with underexposed images, we acquire multiple RAW-format images at each scene's position with $48=(\mbox{6 shutter speeds}\times\mbox{8 ISO settings})$ exposure settings ranging from extreme to mildly underexposure settings. The camera is mounted on a tripod while capturing all the images.
\item We additionally provide long-exposure images, using which as the ground truth, one can evaluate image restoration methods on the task of estimating it from one of the underexposed images. 
\item The current standard for the evaluation of image matching methods is to measure the accuracy of the downstream task, i.e., the estimation of geometric parameters, as was pointed out in recent studies \cite{jin2020image}.
Therefore, we acquire images from two positions to form stereo pairs for each scene along with their ground truth relative pose. To obtain the ground truth pose, we capture a good quality image with a long-exposure setting for each scene position.
\item The dataset contains diverse scenes consisting of 54 outdoor and 54 indoor scenes. 
\end{itemize}

Using this dataset, we experimentally evaluate several existing component methods for the SfM pipeline, i.e., detecting keypoints and extracting descriptors \cite{sift}, finding initial point correspondences, and removing outliers from them \cite{ransac1,brachmann2019neural,superglue}. We choose classical methods and learning-based methods for each. We also evaluate the effectiveness of image enhancement, including classical image-enhancing methods with/without denoising \cite{bm3d}, and a CNN-based method \cite{chen2018learning,zamir2020learning}. The results show the importance of using the RAW-format images instead of using the processed images by the standard RIP. They further provide the strengths and weaknesses of the above component methods, also showing that there is room for further improvement.
\section{Related Work}
\subsection{Matching Multi-view Images} 

Matching multi-view images of a scene is a fundamental task of computer vision, and its research has a long history.
It generally performs the following steps: detecting keypoints/computing local descriptors, establishing initial point correspondences, and removing outliers to finding correct correspondences.
A baseline of this pipeline, built upon traditional methods, consists of
SIFT \cite{sift}, SURF~\cite{surf}, etc.
for detecting interest points and extracting their local descriptor, nearest neighbor search in the descriptor's space for obtaining initial correspondences across images, with an optional `ratio test' step for filtering out unreliable matches \cite{sift}, and RANSAC for outlier removal \cite{ransac1, ransac2}. 

A recent trend is to use CNNs to detect keypoints and/or extract local descriptors. Early studies attempted to learn either keypoint detectors \cite{Webcam, savinov2017quad,barroso2019key}, or descriptors \cite{simo2015discriminative, zagoruyko2015learning,Matchnet,PN-Net,yoo2015multi}. In contrast, in recent studies, researchers have proposed end-to-end pipelines \cite{yi2016lift, superpoint, d2net, lfnet, r2d2, ret3} that can perform the two at once. Despite the success of CNNs in many computer vision tasks, it remains unclear that these learning-based methods have surpassed the classical hand-crafted methods. In parallel to the developments of methods for keypoint detectors and descriptors, several recent studies have developed learning-based methods for initial point matching and outlier removal  \cite{brachmann2019neural, superglue}.

\subsection{Datasets for Image Matching}




There are many datasets created for the research of image matching \cite{oxford,dtu,edge,amos,strecha2008benchmarking, VLBenchmarks, Webcam}. Many recent studies of image matching employ HPatches \cite{hpatches}. 
There are also a number of datasets for visual SLAM and localization/navigation \cite{Aachen1, kitti, robotcar, cmu, SILDa}. 

Some of these datasets provide challenging cases, including illumination changes, matching daylight and nighttime images, motion blur in low-light conditions, etc. However, all these datasets provide only images in the regime where the standard RIP can successfully yield RGB images with a well-balanced brightness histogram. This is also the case with a recent study \cite{jenicek2019no} that analyzes image retrieval under varying illumination conditions. Our dataset contains the images of very dark scenes all in a RAW-format with 14-bit depth. 
In fact, while we have verified the authors' findings in \cite{jenicek2019no} with 8-bit images converted from our RAW-format images using the standard RIP, they do not hold in the case of directly using the RAW-format images, as we will show later.



There are also many evaluation methods for image matching, which are developed aiming at a more precise evaluation \cite{schonberger2017comparative,oxford,dtu,crivellaro2017robust,bian2017gms}. A recent study has introduced a comprehensive benchmark for image matching \cite{jin2020image}. As in this study, the current trend is to focus on the downstream task; the accuracy of the reconstructed camera pose is chosen as a primary metric for evaluation. Following this trend, our dataset provides the ground truth for the relative camera pose between every stereo image pair.

\subsection{Image Enhancement} 

There are many image-enhancing methods that improve the quality of underexposed images. 
Besides basic image processing such as histogram equalization, there are many methods based on different assumptions and physics-based models, etc., such as global analysis and processing based on the inverse dark channel prior \cite{malm2007adaptive, dong2011fast}, the wavelet transform \cite{loza2013automatic}, the Retinex model \cite{park2017low}, and illumination map estimation \cite{guo2016lime}. These methods are proven to be effective for images that are mildly underexposed. 

To deal with more severely underexposed images, Chen \etal proposed a learning-based method that uses a CNN to directly convert a low-light RAW image to a good quality RGB image \cite{chen2018learning}. Creating a dataset containing pairs of underexposed and well-exposed RAW images (i.e., the SID dataset), they train the CNN in a supervised fashion. Their method can handle more severe image noise and color distortion emerging in underexposed images than the previous methods. 
For the problem of enhancing extreme low-light videos, Chen \etal extended this method while creating a dataset for training \cite{chen2019seeing}. In parallel to these studies, Wei \etal have developed a model of image noises, making it possible to synthesize realistic underexposed images \cite{wei2020physics}. They demonstrated that a CNN trained on the synthetic dataset generated by their model performs denoising equally well or even better than a CNN trained on pairs of real under/well-exposed images.

While these studies aim sorely at image enhancement, our study considers the problem of matching images of extremely low-light scenes. Our dataset contains stereo image pairs of multiple scenes; there are 48 low-light RAW images with different exposure settings and one long-exposure reference image for each camera position of each scene. It is noteworthy that they include much more underexposed images than the datasets of \cite{chen2018learning, chen2019seeing}.

\section{Dataset for Low-light Image Matching}

\subsection{Design of the Dataset} \label{overview}

\begin{figure}[t]
\centering
\includegraphics[width=0.95\linewidth]{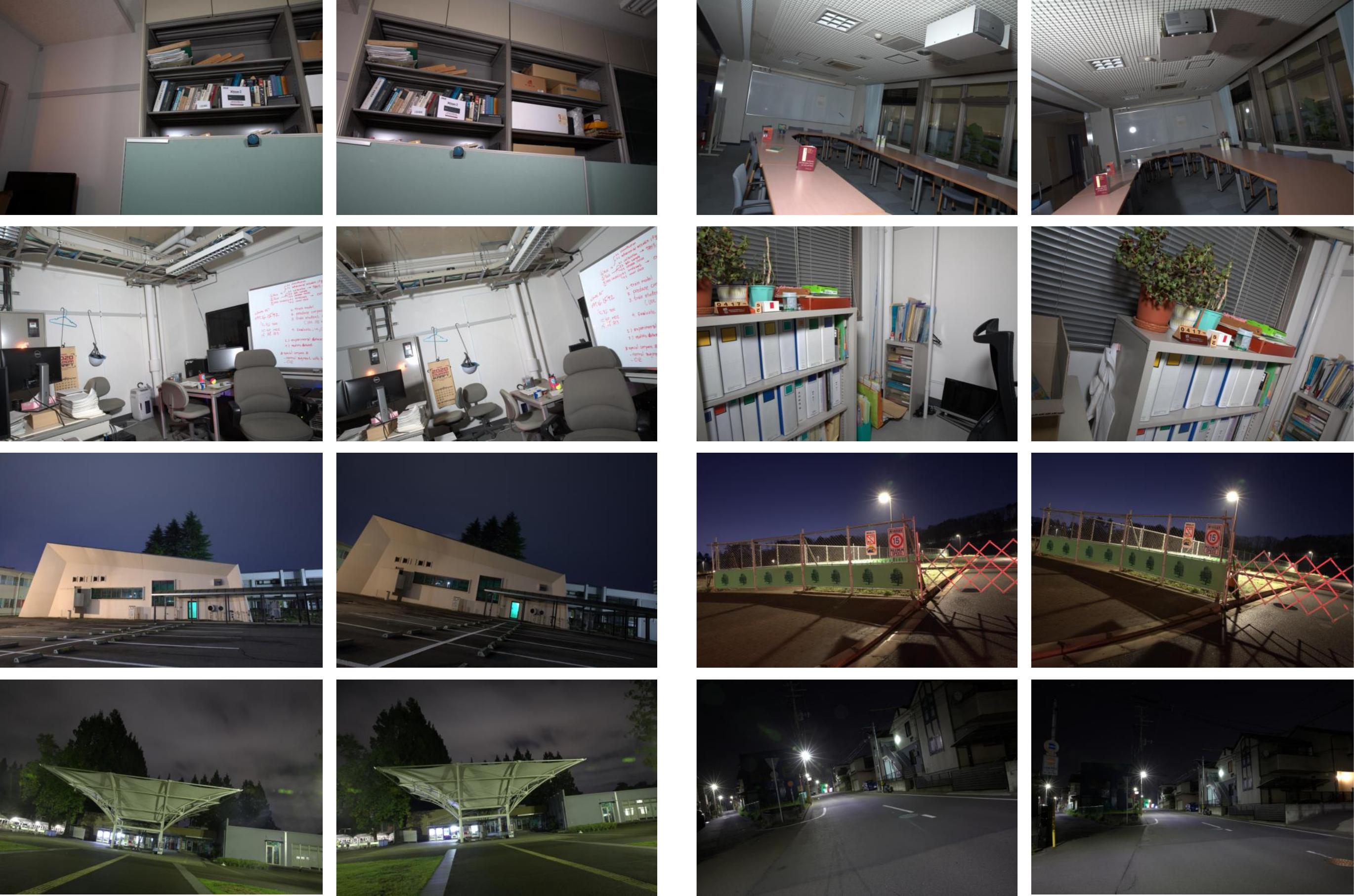}
   \caption{Example stereo image pairs (long exposure versions) of four indoor scenes (upper two rows) and four outdoor scenes (lower two rows). }
\label{fig:references}
\end{figure}

We built a dataset of stereo images of low-light scenes and named it the MID (Matching In the Dark) dataset. It contains stereo image pairs of 54 indoor and 54 outdoor scenes (108 in total). We used a high-end digital camera to capture all the images; they are recorded in a RAW format with 14-bit depths.  Figure~\ref{fig:references} shows example scene images. For each of the 108 scenes, we captured images from two viewpoints with 49 different exposure settings, i.e., 48 exposure settings in a fixed range plus one long exposure setting to acquire a reference image. Note that most of the images are so underexposed that the standard RIP cannot yield reasonable RGB images from them. 

Some of the 48 images of each scene captured with the most underexposed settings are so underexposed that they appear to store only noises; it will be impossible to perform image matching using them, even if we try every one of the currently available methods. Nevertheless, we keep these images in the dataset to assess the lower limit of exposure to which image matching and restoration methods work, not only existing methods but those to be developed in the future. We designed the dataset primarily to evaluate image matching methods in low-light conditions, but the users can also evaluate image-enhancing methods. Our 48 images of each scene contain more severely underexposed ones than any existing datasets for low-light image enhancement (e.g., \cite{chen2018learning}). 

\subsection{Detailed Specifications}

The dataset contains 10,584 ($=108\mbox{(scenes)}\times 2\mbox{(stereo)}\times(48+1)\mbox{(exposure settings)}$) images in total. They are of $6,720\times 4,480$ pixels and in a RAW format of 14 bits per pixel; its Bayer pattern is RGGB. We used Canon EOS 5D Mark IV with a full-frame CMOS sensor and EF24-70mm f/2.8L II USM to capture these images.

For each scene, we set up the camera in two positions to capture stereo images. For each position, we mounted the camera on a sturdy tripod while capturing 49 images. We first captured a long-exposure image, which serves as a reference image; we use it to compute the ground-truth camera poses of the stereo pair, as will be explained in Sec.~\ref{section:gt}. To capture the reference image, we choose exposure time from the range of 10 to 30 seconds, while fixing ISO to 400.

We then captured the low-light images in 48 different exposure settings that are combinations between six exposure times and eight ISO values. The exposure time is chosen from the range of $[1/200,1]$ seconds for the indoor scenes and $[1/200,0.5]$ seconds for the outdoor scenes. The ISO value is chosen from $\{100, 200, 400, 800, 1600, 3200, 6400, 12800\}$. 

The indoor scene images were captured in closed rooms with regular lights turned off; the illuminance at the camera is in the range of 0.02 to 0.3 lux. The outdoor scene images were captured at night under moonlight or street lighting. The illuminance at the camera is in the range of 0.01 to 3 lux.

\subsection{Obtaining Ground Truth Camera Pose} \label{section:gt}


To compare various image matching methods with different local descriptors and keypoint detectors, we need to evaluate the accuracy of the camera poses estimated from their matching results. We consider stereo matching in our dataset, and an image matching method yields the estimate of the relative camera pose between the stereo images. To obtain its ground truth, we use the pairs of the reference images to perform image matching, from which we estimate the relative camera pose for each scene. Following \cite{bian2019evaluation}, we use it as the ground truth after manual inspection along with correction, if necessary, which we will explain later.

The detailed procedure for obtaining the ground truth camera pose for each scene is as follows. 

We first convert the two reference images in the RAW format into RGB space\footnote{Following \cite{chen2018learning}, we used rawpy ({\scriptsize\url{https://pypi.org/project/rawpy/}}), a python wrapper for {\it libraw} that is a raw image processing library (\scriptsize\url{https://www.libraw.org/})}. 
We then covert each RGB image into grayscale and compute keypoints and their descriptors using the difference of Gaussian (DoG) operator and the RootSIFT descriptor \cite{RootSIFT}. We next establish their initial point matches using nearest neighbor search with Lowe's ratio test \cite{sift} with a threshold of $0.8$. 

We then estimate the essential matrix by using the 5-point algorithm with the pretrained neural-guided RANSAC (NG-RANSAC) \cite{brachmann2019neural}. We employed the authors' implementation
for it. We employ NG-RANSAC over conventional RANSAC, since we found in our experiments that it consistently yields more accurate results. Calibrating the camera with the standard method using a planar calibration chart, we decompose the estimated essential matrix and obtain the relative camera pose (i.e., translation and rotation) between the stereo pair. 

As mentioned above, we performed a manual inspection of the estimated essential matrix, ensuring they are reliable enough to be used as the ground truths. We did this by checking if any image point on the paired images satisfies the epipolar constraint given by the estimated essential matrix. To be specific, we manually select a point on either left or right image and draw its epipolar line on the other image.

We then visually check if the corresponding point lies on the epipolar line with a deviation less than one pixel. We chose a variety of points having different depths for this check. If an estimated essential matrix fails to pass this test, we either remove the scene entirely or manually add several point matches to get a more accurate estimate of the essential matrix and perform the above test again. All the scenes in our dataset have passed this test. 

\section{Matching Images in Low-light Scenes}
\label{sec:methods}

This section discusses what methods are applicable to matching low-light images in our dataset. We evaluate those in our experiments. 

\subsection{Conversion of RAW Images to RGB}

\begin{figure*}[t]
\centering
   \includegraphics[width=0.95\textwidth]{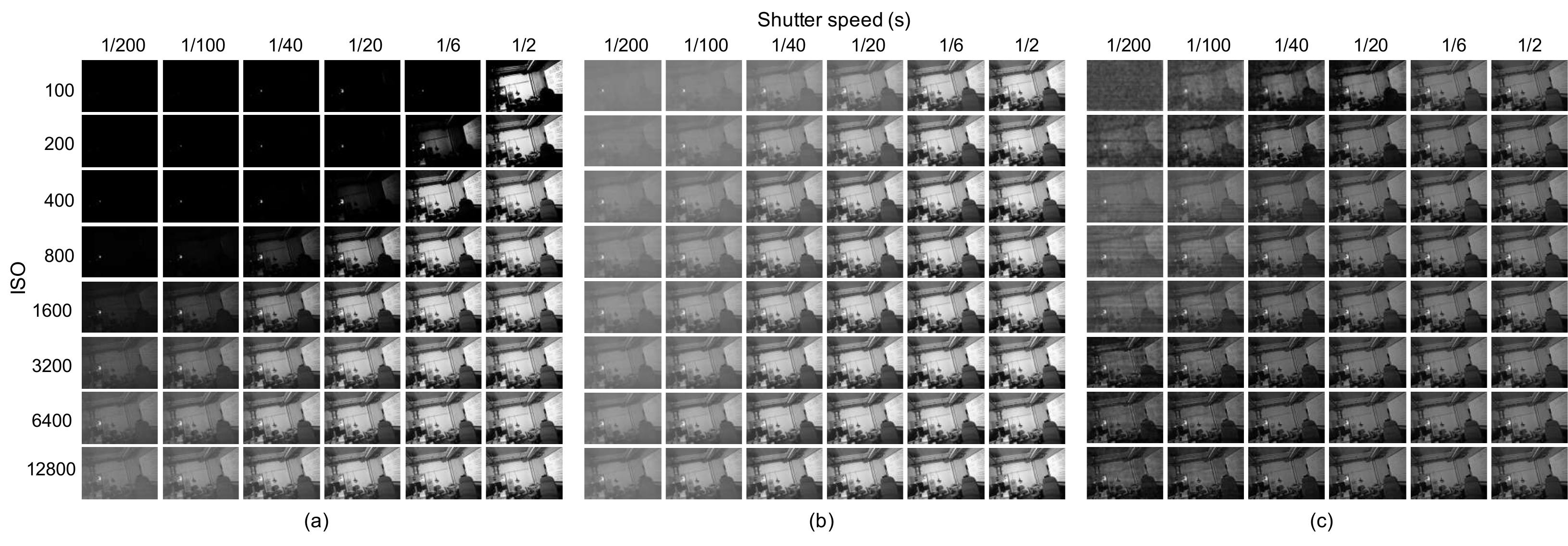}
  \caption{Images of a scene captured from the same camera pose that are converted from their RAW-format originals by three conversion methods. (a) {\bf RIP-HistEq}. 
(b) {\bf Direct-BM3D}. (c) {\bf SID}. See text for these methods. 
  }
  
\label{fig:blackset}
\end{figure*}

As there is currently no image matching method directly applicable to RAW-format images, we consider existing keypoint detectors and local descriptors that receive grayscale images. To cope with the low-light condition, we plug image enhancement methods before keypoint detectors and local descriptors, which we will describe later. 

It is first necessary to convert RAW-format images into RGB/grayscale images. We have two choices here. One is to use the standard RIP that converts RAW to RGB. As mentioned in Sec.~\ref{section:1}, the standard RIP often fails to make use of brightness information stored in the lower bits of RAW signals of dark scenes, due to the requirement for versatility against a variety of scenes with different illumination conditions and also the limit of computational resources available to on-board RIP. To confirm its limitation, we evaluate this standard-camera-pipeline-based conversion in our experiment; we use the LibRaw library using rawpy, a Python image processing module. 

The other choice is to do the conversion without using the standard RIP. We will explain this below, because it is coupled with the image enhancement step.

\begin{figure}[tb]
\centering
   \includegraphics[width=0.95\linewidth]{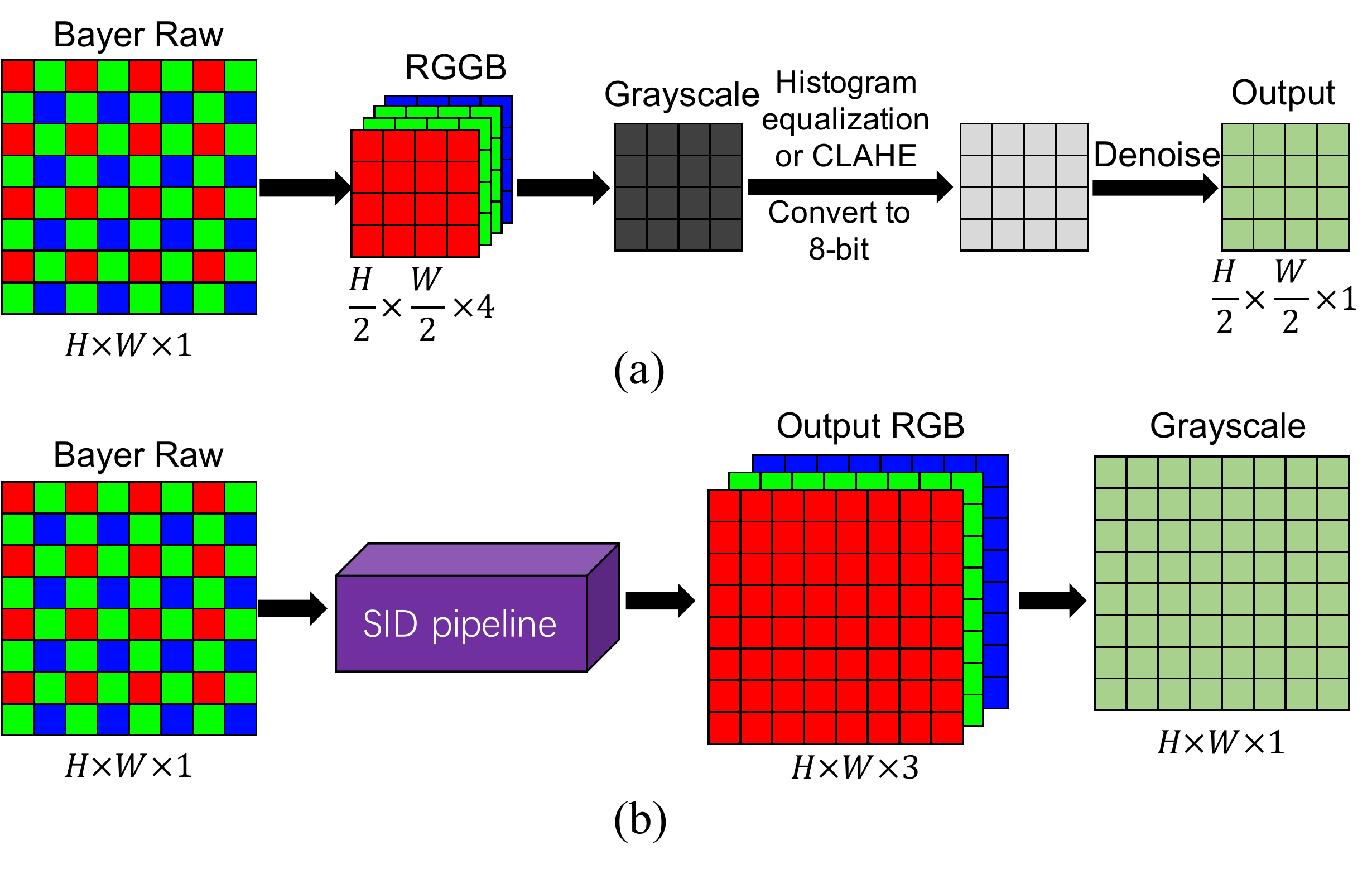}
   \caption{The pipelines of two image-enhancing methods. (a) {\bf Direct-HistEq} or {\bf Direct-CLAHE}. (b) {\bf SID}.}
\label{fig:pipeline}
\end{figure}

\subsection{Image Enhancement}\label{image_enhance}

Thus, we consider two methods, i.e., using the standard RIP for the RAW-to-RGB conversion and directly using  RAW-format images. For each, we consider three different image enhancing methods. 

\subsubsection{Conversion by standard camera pipeline}

When using the standard RIP to convert RAW images, we consider applying the following four methods to its outputs: none, a classical histogram equalization, a contrast limited adaptive histogram equalization (CLAHE), and a CNN-based image enhancement, MIRNet \cite{zamir2020learning}. We choose MIRNet because it is currently the best image-enhancing method applicable to RGB/grayscale images. Figure~\ref{fig:blackset}(a) shows examples of the standard RIP with histogram equalization. We will refer to the four methods as {\bf standard RIP}, {\bf RIP-HistEq}, {\bf RIP-CLAHE}, and {\bf RIP-MIRNet} in Sec.~\ref{sec:exp}.

\subsubsection{Direct Use of RAW-format Images}

We consider two approaches. One is to use standard image processing methods to convert RAW to RGB/grayscale images; see Fig.~\ref{fig:pipeline}(a). For this, we employ the following simple approach. Given a Bayer array containing the input RAW data, we first apply black level subtraction to it and then split the result into four channels; the pixel values are now represented as floating point numbers. We then take the average of the two green channels to obtain an RGB image and convert it to grayscale using the OpenCV function \verb|cvtColor|. 
Next, we perform histogram equalization or CLAHE to improve the brightness of the image. We map the brightness in the range $[m-2d,m+2d]$, where $m$ is the average brightness and $d$ is the mean absolute difference from $m$ to each pixel value, to the range $[0,255]$. Finally, we quantize the pixel depth to 8 bits. We will call this method {\bf Direct-HistEq} or {\bf Direct-CLAHE}. 

We optionally apply denoising to the converted image at the final step. We employ BM3D \cite{bm3d} with a noise PSD ratio of 0.08 in our experiments.
The resulting image will be transferred to the second step of image matching. Figure \ref{fig:blackset}(b) shows examples of the converted images by the method. We will call this method {\bf Direct-BM3D}. 

In parallel to the above, we consider a CNN-based image enhancing method that directly works on RAW-format images; see Fig.~\ref{fig:pipeline}(b).
We employ SID \cite{chen2018learning}, a CNN trained on the task of converting an underexposed RAW image of a low-light scene to a good quality image. It is designed to receive the RAW data of an image and output an RGB image. We calculate the amplification ratio of SID using shutter speed and ISO values between the underexposed and the reference images. As the output of SID is twice as large as others, we downscale the image size by $2:1$ and then convert it into grayscale for image matching; see Fig.~\ref{fig:blackset}(c). We used the pretrained model provided by the authors
, which is trained on the SID dataset. We call this method {\bf SID} in what follows. 


\subsection{Image Matching} 

We consider matching a pair of images of a scene here. It is to establish point correspondences between images while imposing the epipolar constraint on them and estimate the camera pose (\ie., a essential or fundamental matrix) encoded in the constraint. The standard approach to the problem is to first extract keypoints and their local descriptors from each input image, establish initial matching of the keypoints between the images, and finally estimate the camera pose from them. 

There are at least several methods for each of the three steps. There are many classical methods that do not rely on learning data. As with other computer vision problems, neural networks have been applied to each step. They were first applied to the first step, \ie., keypoint detectors \cite{Webcam} and descriptors \cite{Matchnet,simo2015discriminative,PN-Net,yoo2015multi}, to name a few. The next was the third step of robust estimation \cite{moo2018learning,brachmann2019neural}. Recently, SuperGlue \cite{superglue} was proposed, which deals with the step of establishing initial point correspondences.

\section{Experiments}
\label{sec:exp}

We experimentally evaluate the combinations of several methods discussed in Sec.~\ref{sec:methods} using our dataset. 

\subsection{Experimental Configuration}

\subsubsection{Compared Methods}


We choose both classical methods and neural network-based methods for each step of image matching. As for keypoint detection and local descriptors, we choose RootSIFT \cite{RootSIFT} and ORB \cite{orb} as representative classical methods; we consider ORB because it has been widely used for visual SLAM. We use their implementation of OpenCV-3.4.2.
We use SuperPoint
\cite{superpoint}, Reinforced SuperPoint
\cite{bhowmik2020reinforced}, GIFT
\cite{gift}, R2D2
\cite{r2d2}, and RF-Net
\cite{shen2019rf} as representative neural network-based methods.
Furthermore, we employ L2-Net
\cite{L2-net} and SOSNet
\cite{sosnet} as hybrid methods of classical and neural-based methods; they compute local descriptors based on the SIFT keypoints and neural networks. For them, we use the authors' implementation and follow the settings recommended in their papers.

As for outlier removal of point correspondences, we choose RANSAC and NG-RANSAC \cite{brachmann2019neural}. We use the OpenCV-3.4.2 implementation of RANSAC with $\mathtt{threshold}=0.001$, $\mathtt{probability}=0.999$, and $\mathtt{maxIters}=10,000$ with the five point algorithm and use the authors' code for the latter. For obtaining initial point correspondences, we use the nearest neighbor search and also SuperGlue
\cite{superglue}. We apply Lowe's ratio test \cite{sift} with a threshold of $0.8$ to RootSIFT, L2-Net, SOSNet, and RF-Net.



To summarize, we compare the following eleven methods: {\bf SP:} Superpoint + NN + RANSAC, {\bf RSP:} Reinforced SuperPoint + NN + RANSAC, {\bf GIFT:} GIFT + NN + RANSAC, {\bf SP + SG:} SuperPoint + SuperGlue + RANSAC,
{\bf R2D2:} R2D2 + NN + RANSAC, 
{\bf RF:} RF-Net + NN + RANSAC, 
{\bf L2:} L2-Net + NN + RANSAC, 
{\bf SOS:} SOSNet + NN + RANSAC, 
{\bf RS:} RootSIFT + NN + RANSAC, 
{\bf RS + NG:} RootSIFT + NN + NG-RANSAC, {\bf ORB:}
ORB + NN + RANSAC. As for image enhancers, we use the eight methods explained in  Sec.~\ref{image_enhance}., i.e., {\bf standard RIP}, {\bf RIP-HistEq}, {\bf RIP-CLAHE}, {\bf RIP-MIRNet}, {\bf Direct-HistEq}, {\bf Direct-CLAHE}, {\bf Direct-BM3D}, and {\bf SID}. We combine these eight image enhancers with the above eleven image matching methods and evaluate each of the 88 pairs. We resize the output images from each image enhancer to $960\times 640$ pixels and feed it to the image matching step.

\begin{figure}[t]
\centering
   \includegraphics[width=0.95\linewidth]{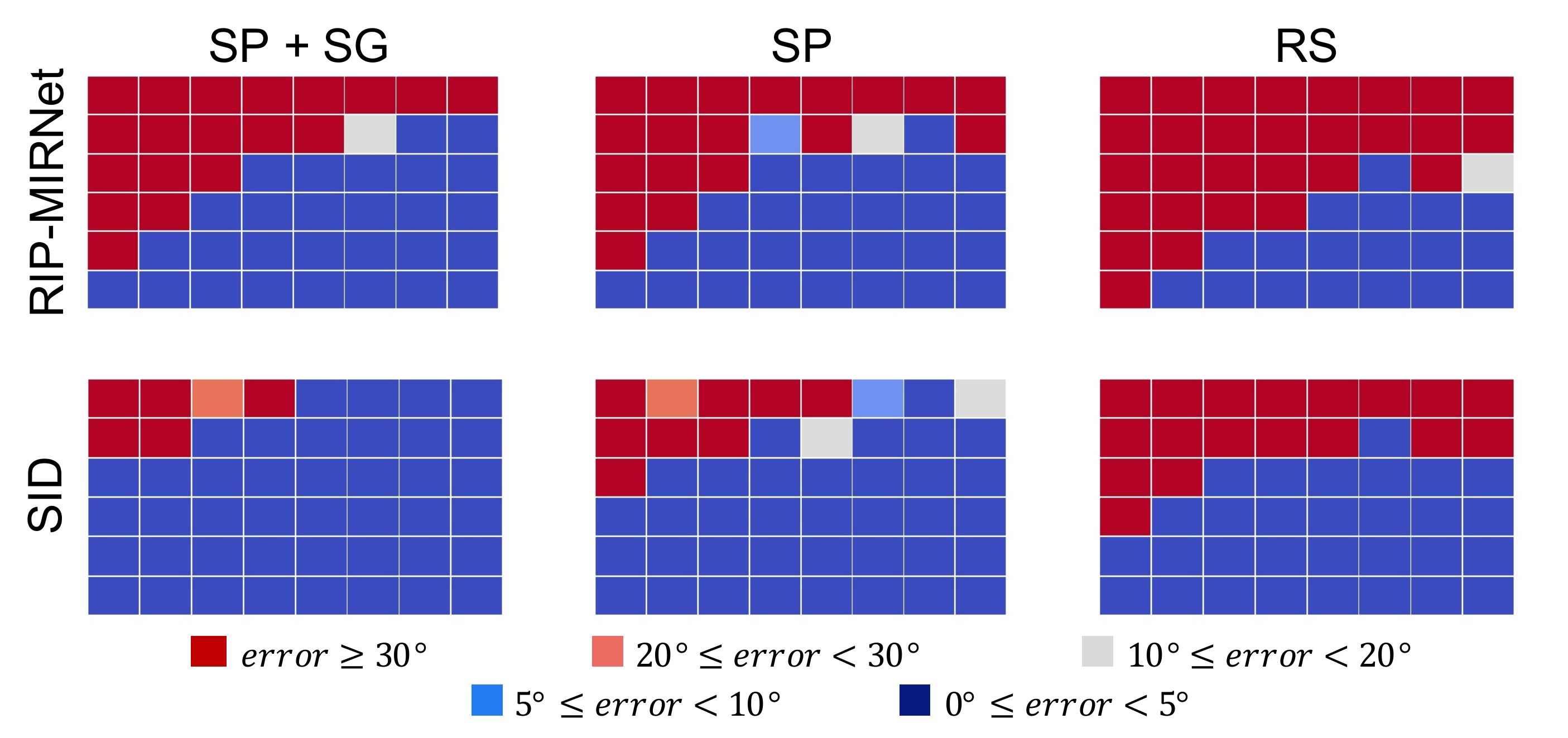}
\caption{Angular errors of the camera pose estimated by several methods for a scene from images with $6\times 8$ different exposure settings. The number of cells with an error lower than a specified threshold quantifies the robustness of the method. }
\label{fig:heat_map}
\end{figure}

\begin{figure*}[t]
\centering
  \includegraphics[width=0.95\textwidth]{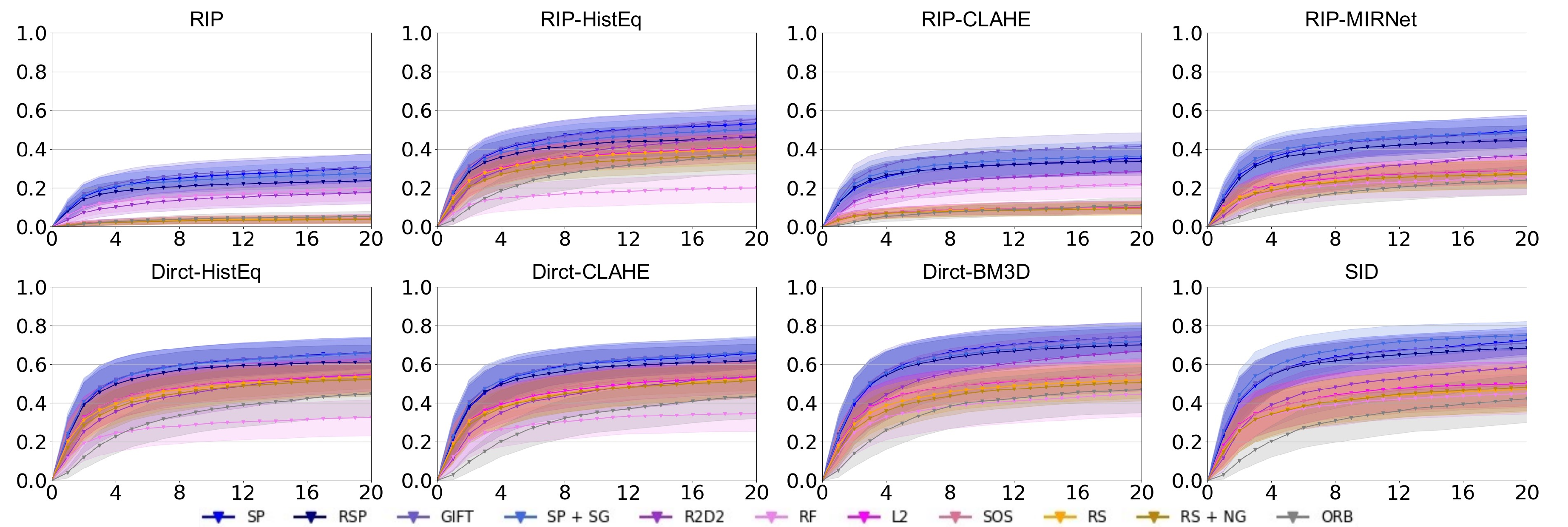}
  \caption{The normalized number $N_\tau$ of the exposure settings  (the vertical axis) for which the estimation error of each method is lower than threshold $\tau$ (the horizontal axis). Each panel shows the means and standard deviations over 54 {\em indoor} scenes for the eleven image matching methods for an image-enhancing method.}
\label{fig:mean-std_in}
\end{figure*}


\subsubsection{Evaluation} 

We compare these methods by evaluating the accuracy of their estimated relative camera pose. We apply each pair of an image enhancer and an image matching method to the stereo images of each scene. We consider only pairs of stereo images with the same exposure setting; there are 48 pairs per each scene. Thus, we have 48 estimates of relative camera pose for each scene.

To evaluate the accuracy of these estimates, we follow the previous work \cite{moo2018learning, brachmann2019neural, superglue}. Specifically, we measure the difference between the rotational component of the ground truth camera pose and its estimate, as well as the angular difference between their translational components. We use the maximum of the two values as the final angular error. Figure \ref{fig:heat_map} shows examples of the results. Each of the colored $6\times 8$ matrices indicate the above angular errors of one of the compared methods for a scene and the 48 exposure settings. 

We are interested in how robust each method will be for underexposed images. To measure this, we count the exposure settings (out of 48) for which each method performs well. To be specific, denoting the above angular error of $i$-th exposure setting by $e_i$ $(i=1,\ldots,48)$, we set a threshold $\tau$ and count the exposure settings with an error lower than $\tau$ as $N_\tau=\sum_{i=1}^{48} 1(e_i < \tau)$, where $1(\mathtt{True})=1$ and $1(\mathtt{False})=0$. We normalize $N_\tau$ dividing by the total number of exposure settings. As shown in Fig.~\ref{fig:heat_map}, the angular error decreases roughly in a monotonic manner from well-exposed toward underexposed settings. Thus, a larger $N_\tau$ means that the method is more robust to underexposure. 

\subsection{Results}

Figure \ref{fig:mean-std_in} 
shows the results for the indoor scenes; see Fig.\,{\color{red}11} in the supplementary for the outdoor scenes.
Table \ref{tab:5deg_mean} shows the mean of $N_\tau$ with $\tau={5^\circ}$ over 54 scenes for indoor and outdoor scenes, 
i.e., the values of the curve in 
Fig.~\ref{fig:mean-std_in} and Fig.\,{\color{red}11}
at the error threshold  $\tau=5^\circ$.
It can be used as a summary of 
Fig.~\ref{fig:mean-std_in} and Fig.\,{\color{red}11}.
We can make the following observations.

\begin{table*}[t]
\centering
\caption{Averaged number $N_\tau$ over 54 scenes of exposure settings for which each method yields a better result than error threshold $=5^\circ$. Extracted from Fig.~\ref{fig:mean-std_in} and Fig.\,{\color{red}11} in the supplementary.
`R-' means `RIP-' and `D-' means `Direct-.'
}
\label{tab:5deg_mean}
\footnotesize
\resizebox{1\linewidth}{!}{
\begin{tabular}{c|c|c|c|c|c|c|c|c||c|c|c|c|c|c|c|c}  
\hline
 \multirow{2}*{}&\multicolumn{8}{c||}{Indoor}&\multicolumn{8}{c}{Outdoor}\\\cline{2-17}
 & RIP & \!R-HistEq\! & \!R-CLAHE\! & \!\!R-MIRNet\!\! & \!D-HistEq\!& \!D-CLAHE\! & \!\!D-BM3D\!\! & ~~SID~~ & RIP & \!R-HistEq\! & \!R-CLAHE\! & \!\!R-MIRNet\!\! & \!D-HistEq\! & \!D-CLAHE\! & \!\!D-BM3D\!\! & ~~SID~~\\
 \hline\hline
 {\bf SP}& 0.223	&	0.421	&	0.275	&	0.381	&	0.548	&	0.540	&	0.596	&	0.583	&	0.233	&	0.379	&	0.269	&	0.352	&	0.460	&	0.475	&	0.502	&	0.500\\\hline
 {\bf RSP} &0.190	&	0.379	&	0.277	&	0.365	&	0.523	&	0.523	&	0.581	&	0.577	&	0.215	&	0.363	&	0.277	&	0.335	&	0.435	&	0.448	&	0.494	&	0.477\\\hline 
  {\bf GIFT} & 	{\bf\underline{0.238}}	&		{\bf\underline{0.427}}	&		{\bf\underline{0.338}}	&	0.390	&		{\bf\underline{0.552}}	&		{\bf\underline{0.550}}	&		{\bf\underline{0.602}}	&	0.583	&	0.254	&	0.375	&	0.321	&	0.358	&	0.475	&	0.477	&	0.506	&	0.492\\\hline 
  {\bf SP + SG}& 0.219	&	0.400	&	0.292	&		{\bf\underline{0.404}} &	0.548	&	0.544	&	0.585	&		{\bf\underline{{\color{red}0.619}}}	&	{\bf\underline{0.302}}	&	{\bf\underline{0.419}}	&	{\bf\underline{0.365}}	&	{\bf\underline{0.410}}	&	{\bf\underline{0.525}}	&	{\bf\underline{0.527}}	&	{\bf\underline{{\color{red}0.577}}}	&	{\bf\underline{0.575}}\\\hline 
  {\bf R2D2} &0.113	&	0.317	&	0.192	&	0.229	&	0.388	&	0.383	&	0.483	&	0.421	&	0.104	&	0.240	&	0.163	&	0.188	&	0.267	&	0.277	&	0.373	&	0.321\\\hline 
  {\bf RF}& 0.138	&	0.154	&	0.152	&	0.192	&	0.256	&	0.275	&	0.346	&	0.358	&	0.160	&	0.146	&	0.183	&	0.202	&	0.225	&	0.244	&	0.323	&	0.325\\\hline 
  {\bf L2} &0.027	&	0.323	&	0.077	&	0.227	&	0.442	&	0.415	&	0.444	&	0.394	&	0.052	&	0.331	&	0.096	&	0.258	&	0.410	&	0.423	&	0.427	&	0.406\\\hline 
  {\bf SOS}& 0.029	&	0.333	&	0.077	&	0.229	&	0.438	&	0.429	&	0.440	&	0.392	&	0.054	&	0.325	&	0.096	&	0.256	&	0.417	&	0.413	&	0.423	&	0.402\\\hline 
  {\bf RS}&	0.025	&	0.317	&	0.071	&	0.210	&	0.423	&	0.404	&	0.410	&	0.369	&	0.046	&	0.317	&	0.094	&	0.242	&	0.410	&	0.413	&	0.404	&	0.406\\\hline 
  {\bf RS + NG}&0.023	&	0.288	&	0.073	&	0.202	&	0.404	&	0.398	&	0.388	&	0.363	&	0.048	&	0.296	&	0.102	&	0.229	&	0.388	&	0.392	&	0.396	&	0.375\\\hline 
  {\bf ORB}& 0.029	&	0.210	&	0.056	&	0.125	&	0.267	&	0.238	&	0.296	&	0.238	&	0.069	&	0.213	&	0.094	&	0.144	&	0.265	&	0.233	&	0.277	&	0.217\\\hline
\end{tabular}}
\end{table*}

\begin{figure*}[t]
\centering
  \includegraphics[width=0.95\textwidth]{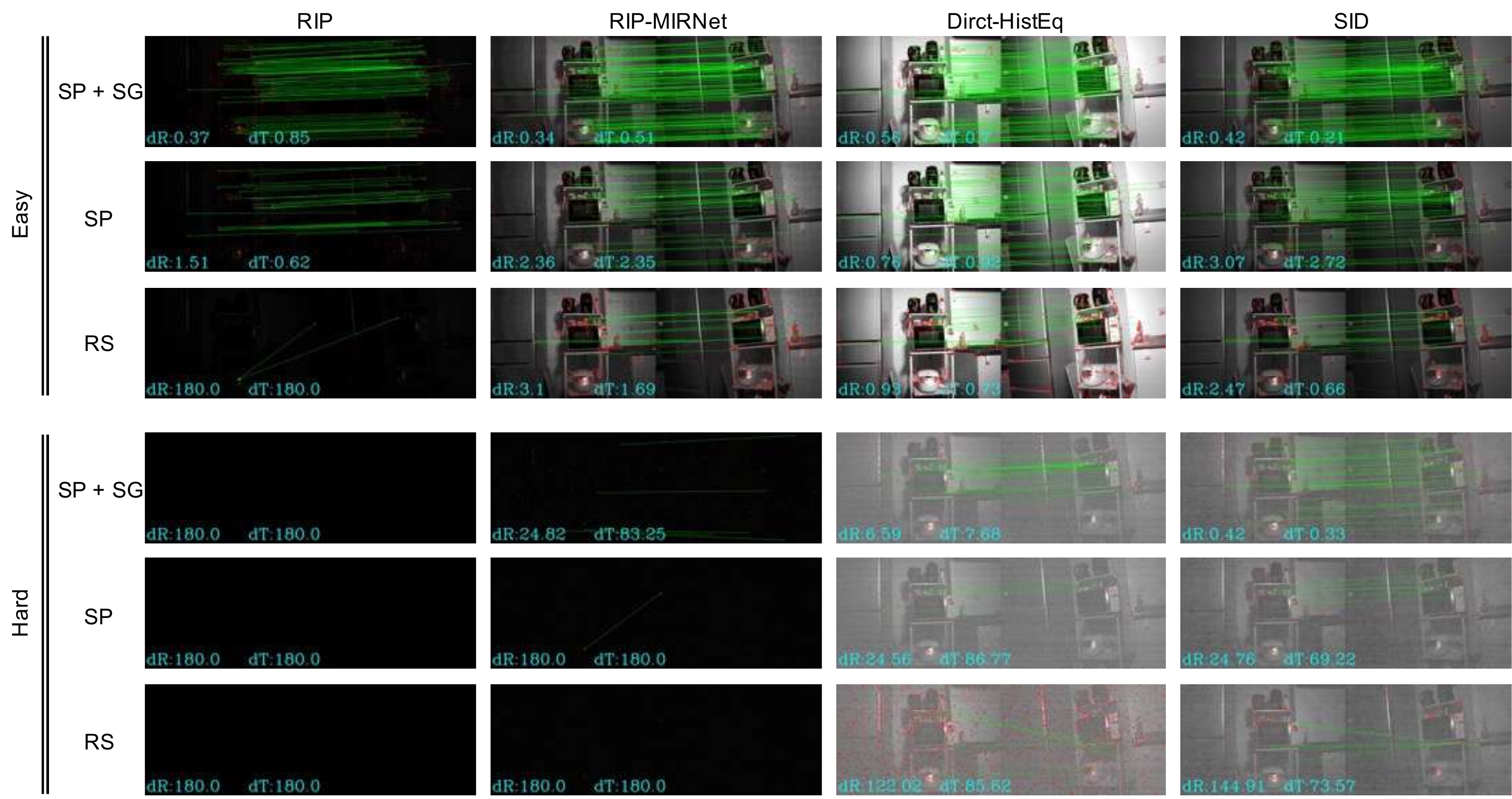}
  \caption{Visualization of the matching results for one of the 54 indoor scenes. Point correspondences judged as inliers are shown in green lines. The combination of three matching methods and the four image enhancing methods are applied to two image pairs with different levels of exposure (i.e., `Easy' and `Hard').}
\label{fig:sample_in}
\end{figure*}

First, the overall comparison of the image enhancers indicates the following: {\em i) Using the standard RIP to convert RAW-format images to 8-bit RGB images before enhancing and matching is inferior to the direct use of RAW-format images.} This shows that the standard RIP cannot utilize the information stored in the low bits of the RAW signals. This fact forms a basis for our dataset. 

Next, the overall comparison of the image matching methods yields the following: {\em ii)  SP and its variants are clearly better than the other methods.} For example, SP and GIFT outperform RS and R2D2 in all cases. This may somewhat contradict previous reports \cite{bhowmik2020reinforced,jin2020image} that while SP is superior to SIFT in the homography-based evaluation using the HPatches dataset, the superiority is not observed in the evaluation with non-planar scene matching.
Additionally, {\em iii)  SP+SG performs the best in many cases.} However, the gap to other methods considerably differs between the indoor and the outdoor scenes. For the outdoor scenes, the gap to the second-best methods tends to be large, whereas, for the indoor scenes, it is not so large. 


The comparison within standard-camera-pipeline-based enhancers indicates the following. {\em iv) The results of the standard RIP (without any enhancement) are the worst. Comparing RIP-HistEq and RIP-MIRNet, the former is comparable or even better than the latter.} This agrees with the results reported in the recent study of Jenicek and Chum \cite{jenicek2019no}, where the authors use 8-bit RGB images outputted from the standard RIP. 

Finally, the comparison within the enhancers using RAW-format images shows the following. 
{\em v) For the outdoor scenes, the four enhancers show similar performance in many cases. When used with SP+SG, both BM3D and SID perform better than Direct-HistEq and Direct-CLAHE; the two show the best performance.
For the indoor scenes, while there is a similar tendency, SID shows a good margin with others only when used with SP+SG.}
It is noteworthy that the superiority of SG depends on the chosen image enhancer, regardless of whether they are applied to indoor or outdoor scenes; this tendency cannot be predicted sorely from the performance of SP. 

We conclude that if we use SG, we should choose SID for the image enhancer, which achieves the best performance; if we do not, we should use BM3D since it achieves good performance overall.
This conclusion differs from that with the standard-camera-pipeline-based enhancers (i.e., (iv)), which is another evidence that the proposed dataset offers what is unavailable in the previous datasets providing only low-bit depth images. Figure \ref{fig:sample_in} shows the visualization of a few matching results for an indoor scene.




\section{Summary and Discussion}


This paper has presented a dataset created for evaluating image matching methods for low-light scene images. It contains stereo images of diverse low-light scenes (54 indoor and 54 outdoor scenes). They are captured with 48 different exposure settings, including from mildly to severely underexposed ones. The dataset provides ground truth camera poses to evaluate image matching methods in terms of the accuracy of estimated camera poses. 

We have reported the experiments we conducted to test multiple combinations of existing image-enhancing methods and image-matching methods. The results can be summarized as follows. 
\begin{itemize}
    \item The direct use of the RAW-format images shows a clear advantage over the standard RIP. Using the standard RIP yields only suboptimal performance, as it cannot utilize information stored in the lower bits of RAW-format signals. Moreover, when using the standard RIP, using classical histogram equalization or the state-of-the-art CNN-based image-enhancing method does not make a big difference, as reported in \cite{jenicek2019no}. 
    \item SuperPoint and its variants work consistently better than RootSIFT. 
    \item SID is the best image enhancer when using SuperPoint+SuperGlue. Otherwise, BM3D and SID perform equally well and better than the sole use of histogram equalization.
\end{itemize}


While the above is our conclusion about the combinations of currently available methods, we think there remains much room for improvement.
For instance, we manually chose the range of 14-bit RAW signal and converted it into 8-bit images, and applied Superpoint to them.
It is observed that the manual method yields significantly better results than the image enhancers tested in this paper, showing that none of the tested methods can choose the best range in the 14-bit RAW signals for image matching; see Sec.\,{\color{red}B} in the supplementary for details.
The standard image enhancers are designed to yield images that appear the most natural, which should differ from the best image for image matching. We will explore this possibility in a future study.

\bigskip
\noindent{\bf Acknowledgments: } This work was partly supported by JSPS KAKENHI Grant Number 20H05952 and JP19H01110.

{\small
\bibliographystyle{ieee_fullname}
\bibliography{egbib}
}

\appendix
\section*{Appendix}

\section{Distinction from Existing Datasets}
\begin{figure}[t]
\centering
  \includegraphics[width=1\linewidth]{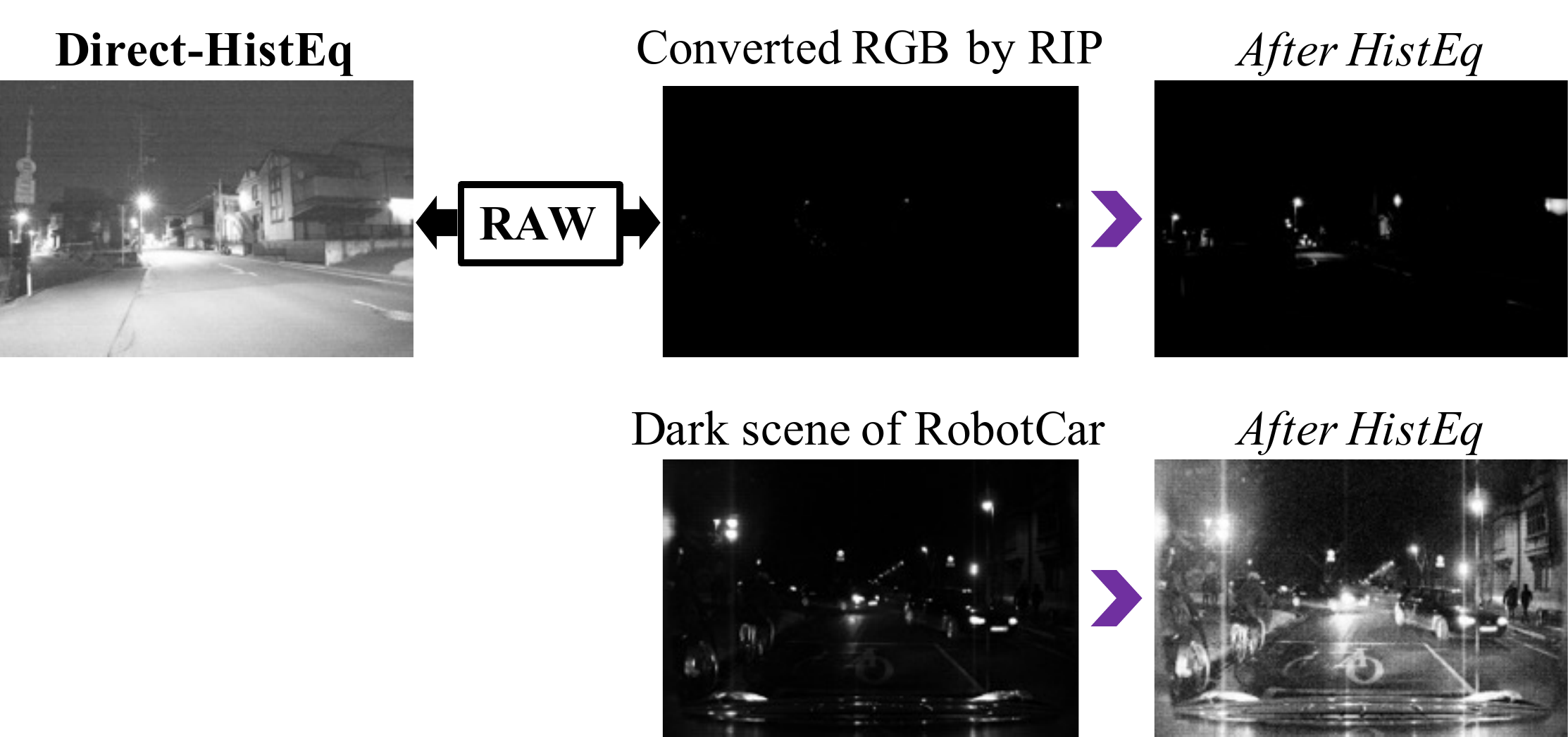}
  \caption{Comparison between RobotCar\,[{\color{green}33}] and our dataset.}
\label{fig:RobotCar_vs_MID}
\end{figure}

Figure \ref{fig:RobotCar_vs_MID} shows example images of our dataset and RobotCar\,[{\color{green}33}]. Most of our images are darker than the darkest one of RobotCar.  The standard raw image processing yields mostly black images from them. Nevertheless, one can derive sufficient info from their RAW signals, when treating them properly.

\section{Performance Comparison to a Manual Adjustment}
\begin{figure}[t]
\centering
  \includegraphics[width=1\linewidth]{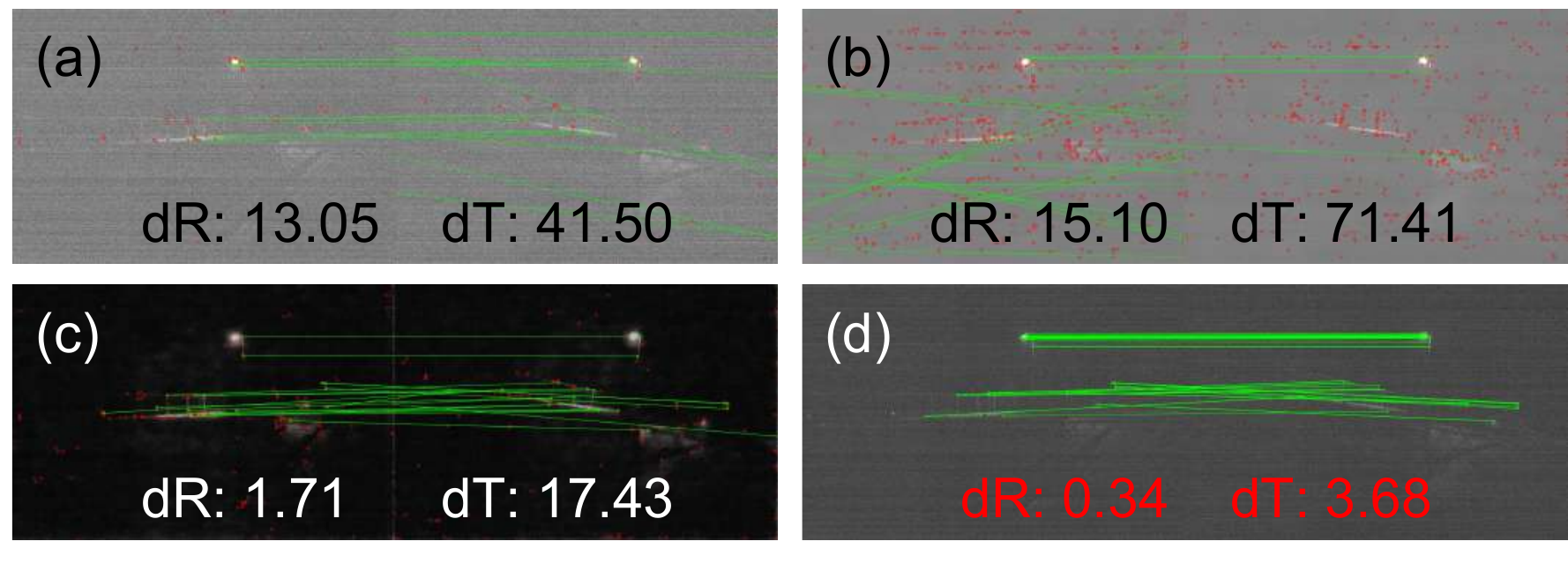}
  \caption{Matching results of {\bf SP} with (a) {\bf Direct-HistEq}, (b) {\bf Direct-BM3D}, (c) {\bf SID}, and (d) Images obtained by manual adjustment of brightness range in 14-bits RAW signals.}
\label{fig:black_sample}
\end{figure}

Figure \ref{fig:black_sample} shows the results of using SuperPoint with the three image-enhancing methods and the result obtained by using SuperPoint on a manually converted 8-bit image from the same RAW image. To be specific, we manually chose the range of 14-bit RAW signal and converted it into an 8-bit image. The values of `dR' and `dT' indicate the rotation and translation errors for each method. It is observed that the manual method yields significantly better results than others and indicates that there is stil much room for improvement.

\section{All Samples of Scene Images in the Dataset}

\begin{figure*}[t]
\centering
   \includegraphics[width=1\textwidth]{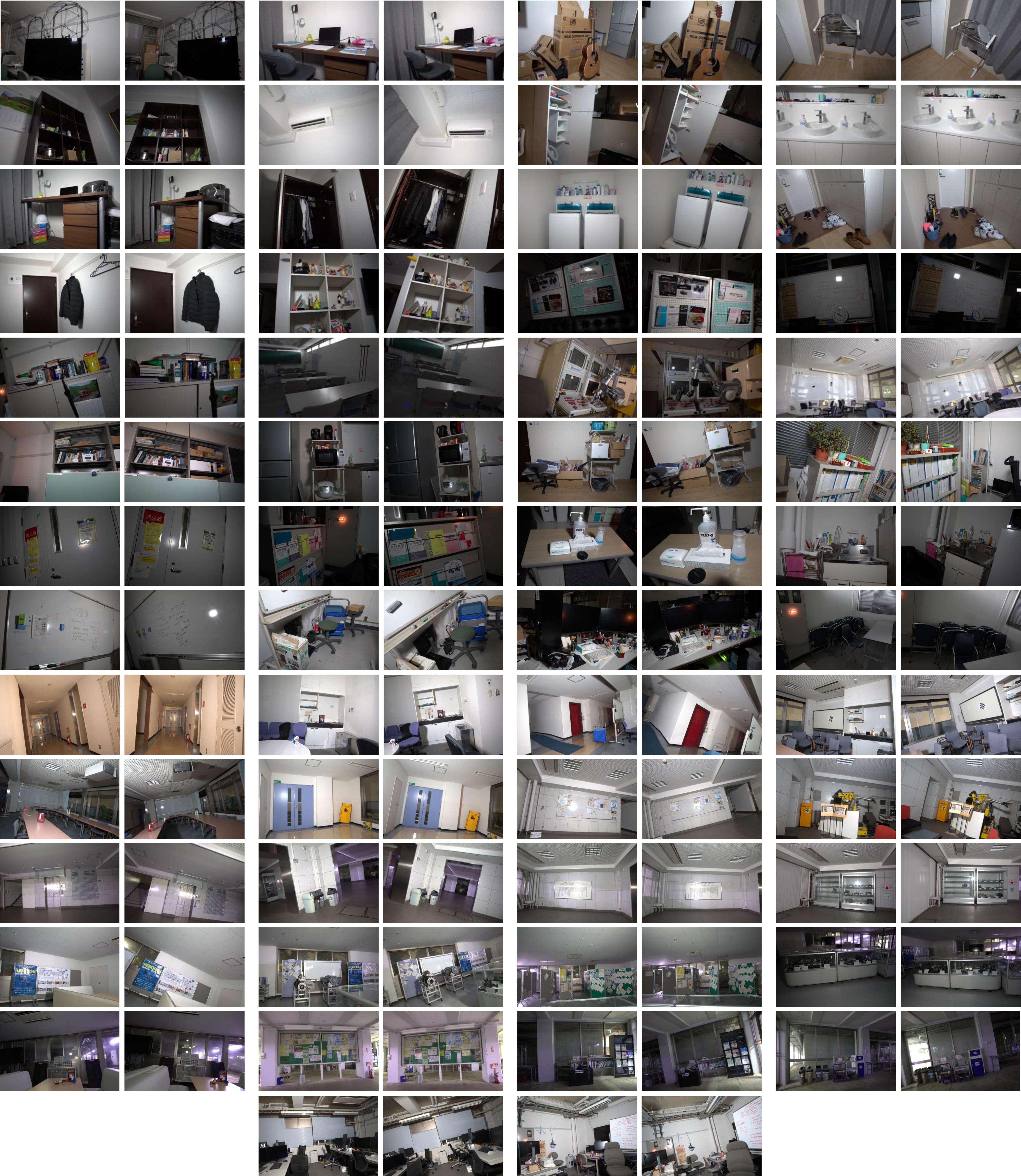}
  \caption{Samples of all image pairs (long  exposure  versions) of the indoor scenes.}
\label{fig:extra_data_in}
\end{figure*}

\begin{figure*}[t]
\centering
   \includegraphics[width=1\textwidth]{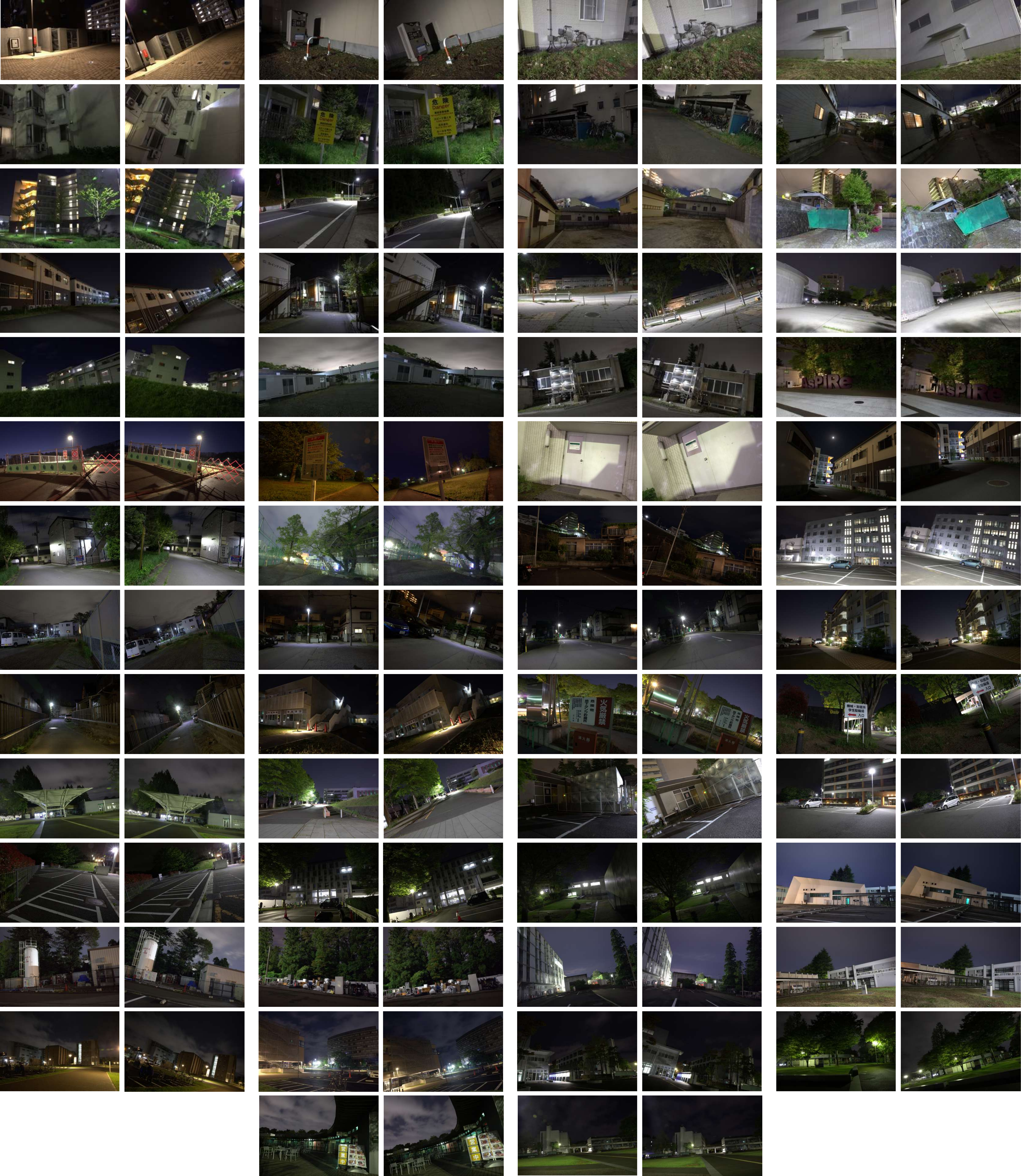}
  \caption{Samples of all image pairs (long  exposure  versions) of the outdoor scenes.}
\label{fig:extra_data_out}
\end{figure*}

Figures \ref{fig:extra_data_in} and \ref{fig:extra_data_out} show all samples of indoor and outdoor scenes in our dataset, respectively.  All images are obtained from the long exposure RAW-format images by the standard RIP.

\section{More Results of Image Matching}\label{E_Evaluation}

\begin{figure*}[t]
\centering
  \includegraphics[width=1\textwidth]{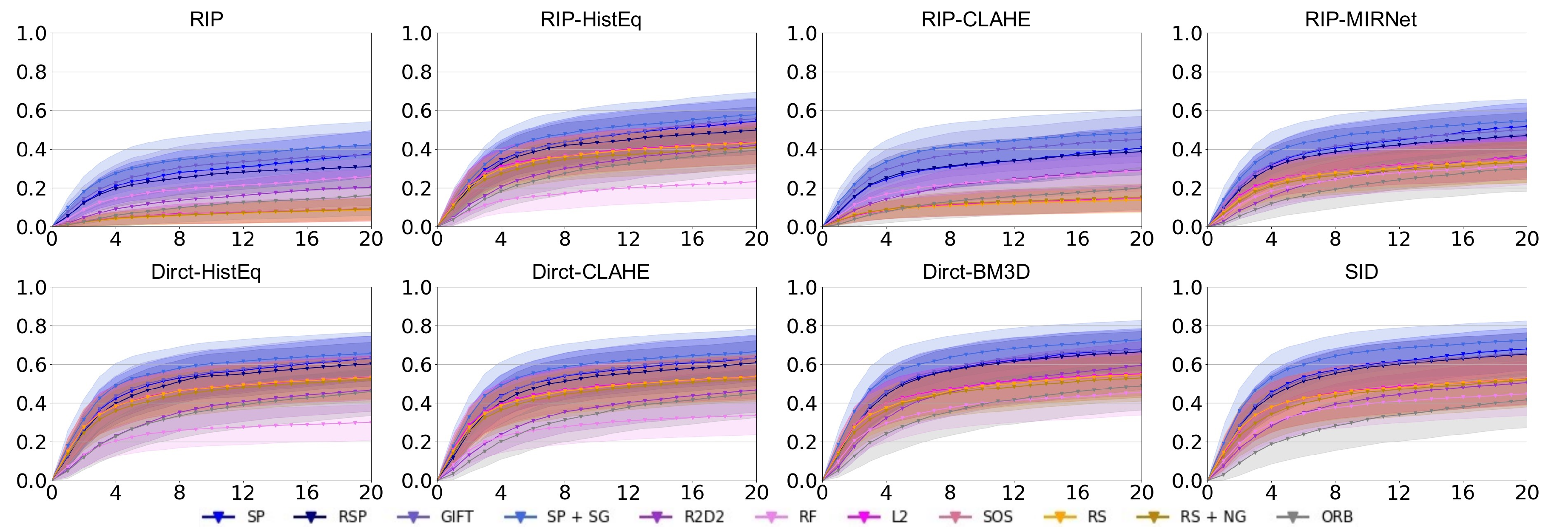}
  \caption{The normalized number $N_\tau$ of the exposure settings  (the vertical axis) for which the estimation error of each method is lower than threshold $\tau$ (the horizontal axis). Each panel shows the means and standard deviations over 54 {\em outdoor} scenes for the eleven image matching methods for an image-enhancing method.}
\label{fig:mean-std_out_al}
\end{figure*}

Figure \ref{fig:mean-std_out_al} shows the normalized number $N_\tau$ of the exposure settings for which the estimation error is lower than threshold $\tau$ averaged over 54 {\em outdoor} scenes.

\begin{figure*}[t]
\centering
   \includegraphics[width=1\textwidth]{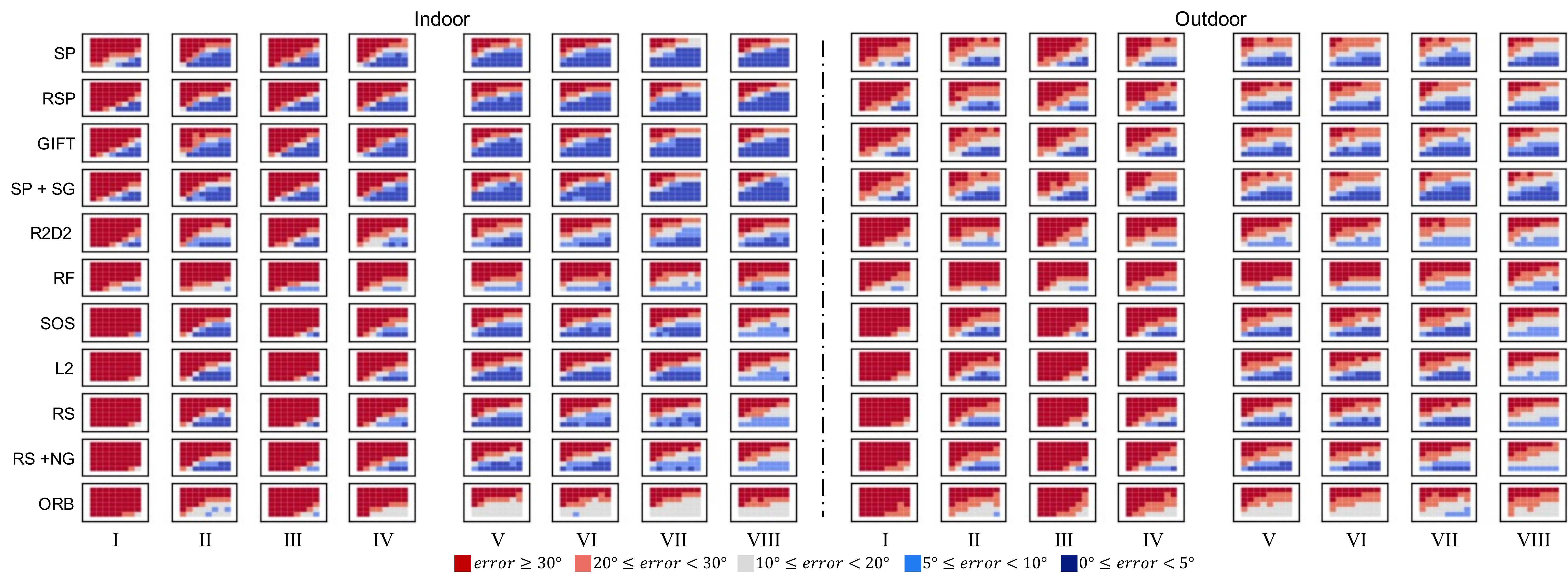}
  \caption{Average angular errors of the camera pose estimated by the 88 methods (i.e., eight image enhancers with eleven image matching methods) over all the 54 scenes for each of the $6\times 8$ exposure settings.  (I) {\bf RIP}. (II) {\bf RIP-HistEq}. (III) {\bf RIP-CLAHE}. (IV) {\bf RIP-MIRNet}. (V) {\bf Direct-HistEq}.  (VI)  {\bf Direct-CLAHE}. (VII) {\bf Direct-BM3D}. (VIII) {\bf SID}.}
\label{fig:heat_map_final}
\end{figure*}

Figure \ref{fig:heat_map_final} shows the average angular errors of the camera pose estimated by the compared 88 methods (i.e., eight image enhancers with eleven image matching methods) over all scenes for each of the $6\times 8$ exposure settings.

\section{Visualization of Matching Results}
\begin{figure*}[t]
\begin{center}
   \includegraphics[width=1\textwidth]{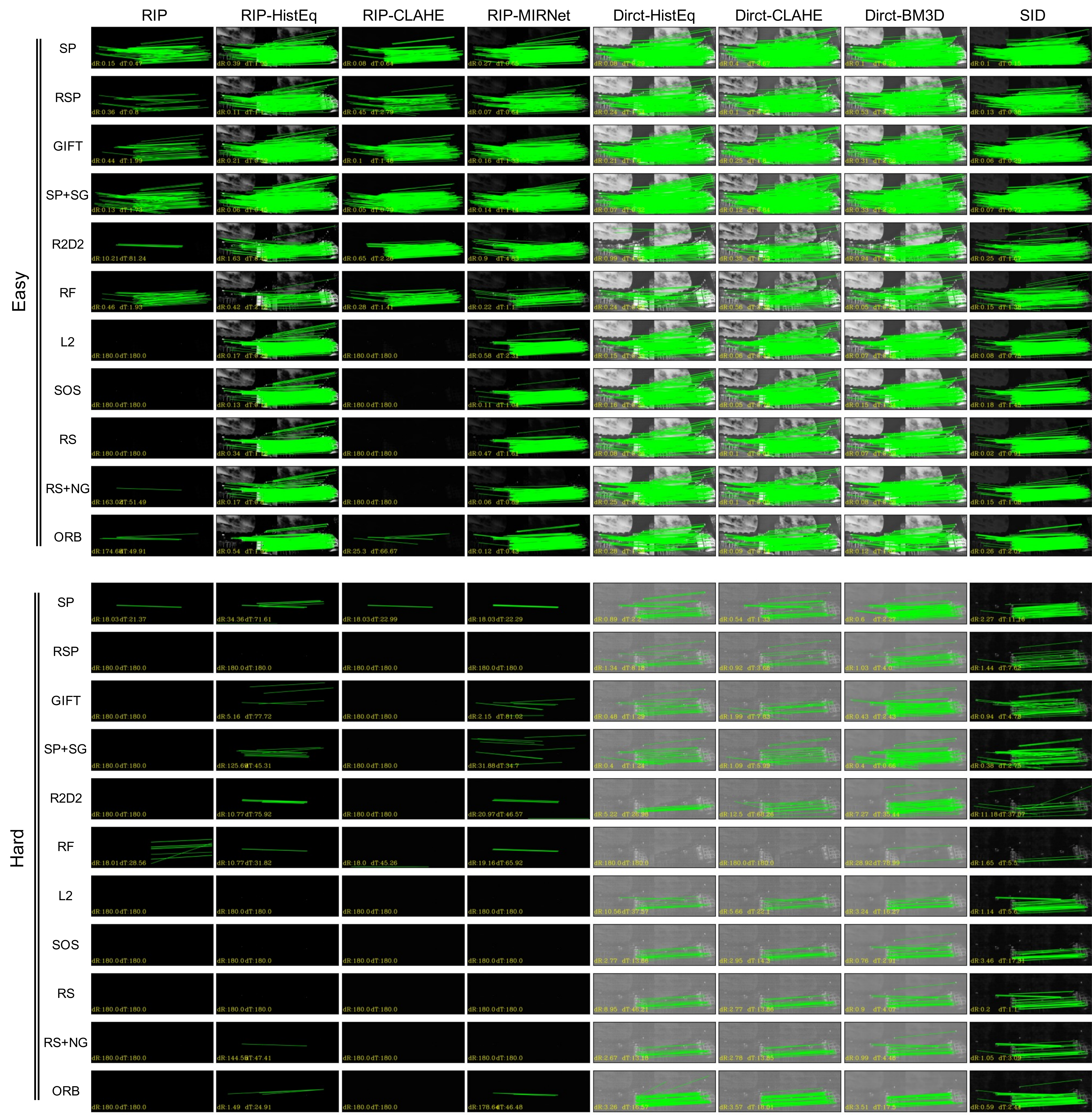}
\end{center}
  \caption{Visualization of the matching results for one of the 54 outdoor scenes. Point correspondences judged as inliers are shown in green lines. The combination of eleven matching methods and the eight image enhancing methods are applied to two image pairs with different levels of exposure (i.e., `Easy' and `Hard').}
\label{fig:sample_out}
\end{figure*}

Figure \ref{fig:sample_in} and \ref{fig:sample_out} show examples of the visualization of the matching results by the 88 methods for an indoor and an outdoor scene, respectively.

\begin{figure*}[t]
\begin{center}
   \includegraphics[width=1\textwidth]{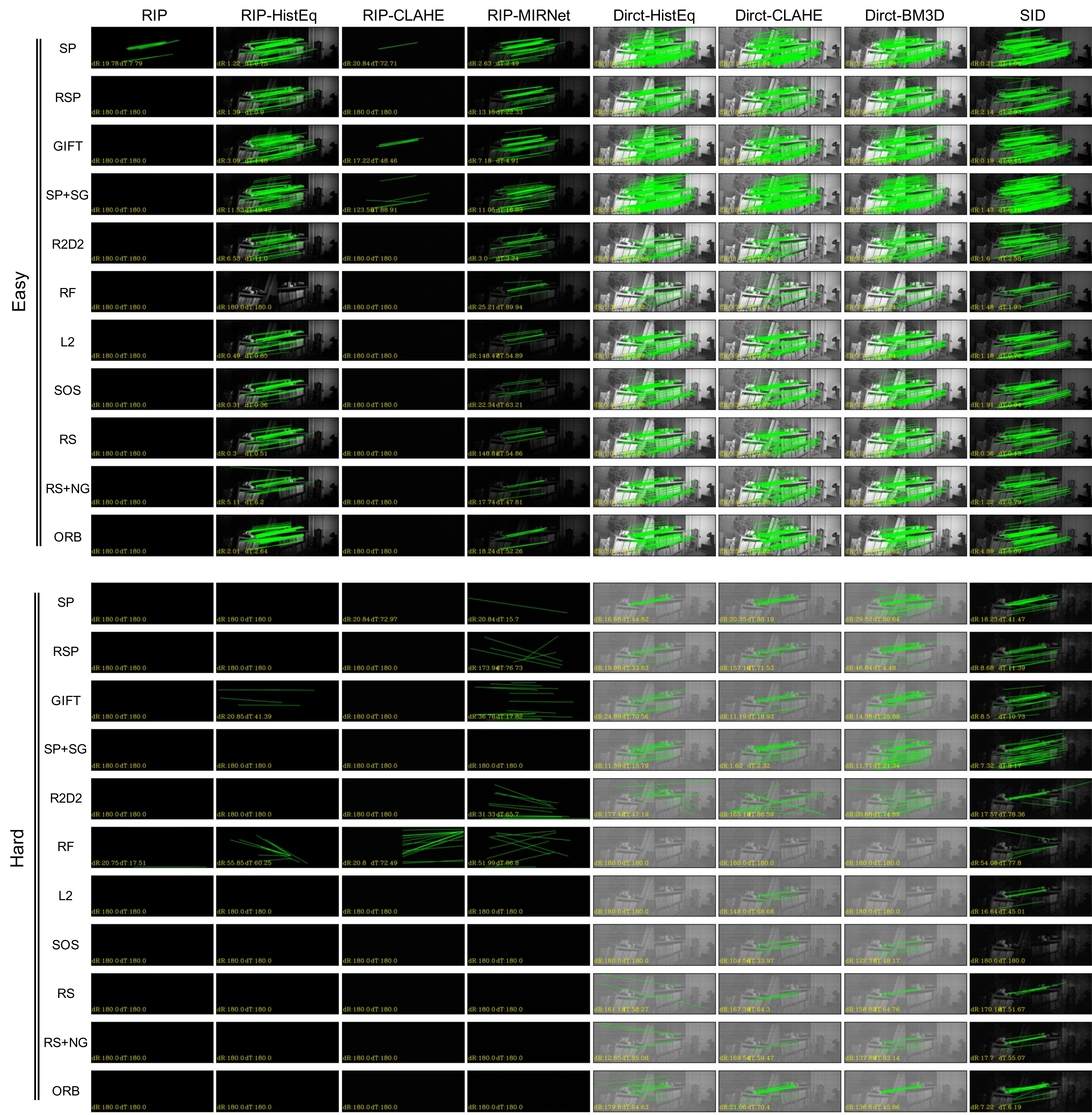}
\end{center}
  \caption{Visualization of the matching results for one of the 54 indoor scenes. Point correspondences judged as inliers are shown in green lines. The combination of eleven matching methods and the eight image enhancing methods are applied to two image pairs with different levels of exposure (i.e., `Easy' and `Hard').}
\label{fig:sample_in}
\end{figure*}

\end{document}